# DEVELOPMENT OF AN NLP-DRIVEN COMPUTER-BASED TEST GUIDE FOR VISUALLY IMPAIRED STUDENTS

Tubo, Faustinah Nemieboka [1], Ikechukwu E. Onyenwe[2], and Doris C. Asogwa[3]
[1]Department of Computer Science, Nnamdi Azikiwe University Awka, Anambra State, Nigeria
[1]faustinah..tubo1@gmail.com, [2]ie.onyenwe@unizik.edu.ng and [3]dc.asogwa@unizik.edu.ng

**Abstract -** In recent years, advancements in Natural Language Processing (NLP) techniques have revolutionized the field of accessibility and exclusivity of testing, particularly for visually impaired students (VIS). CBT has shown in years back its relevance in terms of administering exams electronically, making the test process easier, providing quicker and more accurate results, and offering greater flexibility and accessibility for candidates. Yet, its relevance was not felt by the visually impaired students as they cannot access printed documents. Hence, in this paper, we present an NLP-driven Computer-Based Test guide for visually impaired students. It employs a speech technology pre-trained methods to provide real-time assistance and support to visually impaired students. The system utilizes NLP technologies to convert the text-based questions and the associated options in a machine-readable format. Subsequently, the speech technology pre-trained model processes the converted text enabling the VIS to comprehend and analyze the content. Furthermore, we validated that this pre-trained model is not perverse by testing for accuracy using sample audio datasets labels (A, B, C, D, E, F, G) to compare with the voice recordings obtained from 20 VIS which is been predicted by the system to attain values for precision, recall, and F1-scores. These metrics are used to assess the performance of the pre-trained model and have indicated that it is proficient enough to give its better performance to the evaluated system. The methodology adopted for this system is Object Oriented Analysis and Design Methodology (OOADM) where Objects are discussed and built by modeling real-world instances.

## 1. Introduction

Computer-Based Test (CBT) is a method of conducting assessments, exams, or tests using a computer and digital technology. The idea of CBT shows its efficient test administration, faster result processing, reduced human error in scoring, and adaptability in presenting questions based on a test-taker's performance. It is widely used in education, recruitment, and professional certification processes due to its convenience and accuracy. Despite its usefulness in the education sector, it is not fair to visually impaired students in terms of accessibility and effective use of the application. In order to accommodate visually impaired students, NLP in years have been helpful to visually impaired students in terms of accessibility to printed documents. By leveraging NLP's capabilities, educators, and developers can create inclusive learning environments that cater for the diverse needs of all learners. Through text-to-speech, audio description, language assistance, and other NLP applications, visually impaired students can gain equal access to information, participate actively in academic pursuits, and embark on a journey of knowledge and self-discovery. As technology continues to evolve, the future for visually impaired students is brighter than ever. NLP combines the field of linguistics and computer science to decipher language structure and guidelines and to make models that can comprehend, break down, and separate significant details from text and speech. In the context of speech processing, NLP involves processing spoken language by converting it into text that can be analyzed and manipulated by a computer. This is typically done through a process called automatic speech recognition (ASR), which involves using statistical models and machine learning algorithms to transcribe spoken words into text. Once the spoken language is transcribed into text, NLP techniques is then used to analyze the text and extract useful information such as the sentiment of the speaker, the topics being discussed, and the intent behind the speech. This information can then be used to develop applications such as voice assistants, speech-to-text tools, and sentiment analysis tools. The key benefits of NLP for visually impaired students are Text-to-Speech (TTS) and Speech-to-Text (STT), Audio Description, Language Assistance and Learning Support, Text Summarization and Simplification, Language Translation, and Navigational Support. With the transforming power of NLP in CBT, visually impaired students can actively participate in CBT exams and malpractices will be minimal, especially from the volunteer who renders help to the visually impaired student during the exercise. Other parts of the paper include the literature review, methodology used, result, and conclusion.

## 2. RELATED WORKS

Kuyoro et al., (2016) designed and implemented a Computer-Based Testing System (CBT) to reduce the delay in the notification of a student's final examination scores, tests, and assignments. The COMPUTER-BASED TEST was developed and designed using the waterfall model of the Software Development Life Cycle (SDLC). Implementation was done using open-source web-based technologies such as MySQL, PHP, Javascript, Cascading Style Sheet (CSS), and Hypertext Markup Language (HTML). The system consists of four modules, the index module, the admin module, the lecturer module, and the student's module. Bajeh & Mustapha, (2012) identified the need for improving the current CBT software systems employed by tertiary institutions. The proposed features for improving the reviewed systems include state persistency (in order to prevent data loss in case of any interruption or disruptions that may occur during examinations) and Question Categorization (in order to fairly evaluate candidates by posing questions with overall equal strength to each and every candidate taking an examination). Michael & Berry (2013) developed an Audio assistive technology and testing accommodations which has become an increasingly prevalent and potentially useful means of promoting inclusivity in education. Technologies such as text-to-speech and other forms of audio information representation have helped to make curricula more accessible to people with visual impairments and other disabilities. Auditory accommodations in educational testing have also been implemented in an attempt to ensure equitable access to educational evaluations for people with disabilities. Afolabi, (2014) designed and implemented a Computer-Based Testing software using a speech recognition system. The Computer-Based Test Speech software is an application that supports e-learning via the user's speech. The speech SDK (Software Development Kits) used has an inbuilt speech recognition engine and an inbuilt simulator. The engine provides a speech recognizer that can recognize the user's speech and integrate it with the built-in simulator. Sadaf et al., (2016) focused on developing an interactive framework where speech capabilities can be applied to it. The basic purpose of the framework is to develop independent voice-enabled applications, which can work on speech input and output. The interactive framework can be used by any audience or people with any impairment by enabling them to use the application easily to perform certain tasks on the computer using voice commands. In order to maintain the performance of their system they used a recognition engine to have the effective response of the system. The APIs have been studied carefully and the selection is based on the recognition rate and availability factor. The API chosen here is SAPI using C sharp. Vats, Tandon, & Sinha (2016) developed a voice-based application named "Examination Portal for Blind Person." This system uses a Speech Recognition System (SRS) for converting voice-based answers into text format. The system can handle MCQ-type questions only for the examination. The system can only recognize UK English and it needs a noiseless environment to work effectively. However, this system has no methods for automatic evaluation of answers. Onyesolu & Chimaobi (2017) designed and developed a Computer-Based Testing system with voice command which was proposed to be used in Joint Admission and Matriculation Board examinations. The system reads onscreen questions to candidates, using the earpiece, and allows candidates most especially the physically challenged to make input or control the system using voice. The system is made up of various forms and modules that make up the complete Computer-Based Test system. The suggested system "JAMB Computer-Based Test with the voice command" allows students to not only hear the question straight from the computer but also to answer the questions using their voice. Nwafor & Onyenwe (2020) developed an NLP-based system for automatic MCQG for Computer-Based Testing Examination (CBTE). The research work achieved an automatic generation of correct and relevant questions from textual data. They used the NLP technique to extract keywords that are important words in a given lesson material. This is used to process the lesson materials/documents fed by the teacher into multi-choice questions alongside the answers to each question. The NLP processes are applied using TFIDF and N-gram. This system only applies to the text of the document or lesson material in extracting keywords presented by the teacher. The document is converted into a text file, loaded into the system memory, and stored on the system. The text is split into sentences. The split sentences are tokenized, from which the corpus is built in TF-IDF and N-gram mode.

3. **Research Methodology**

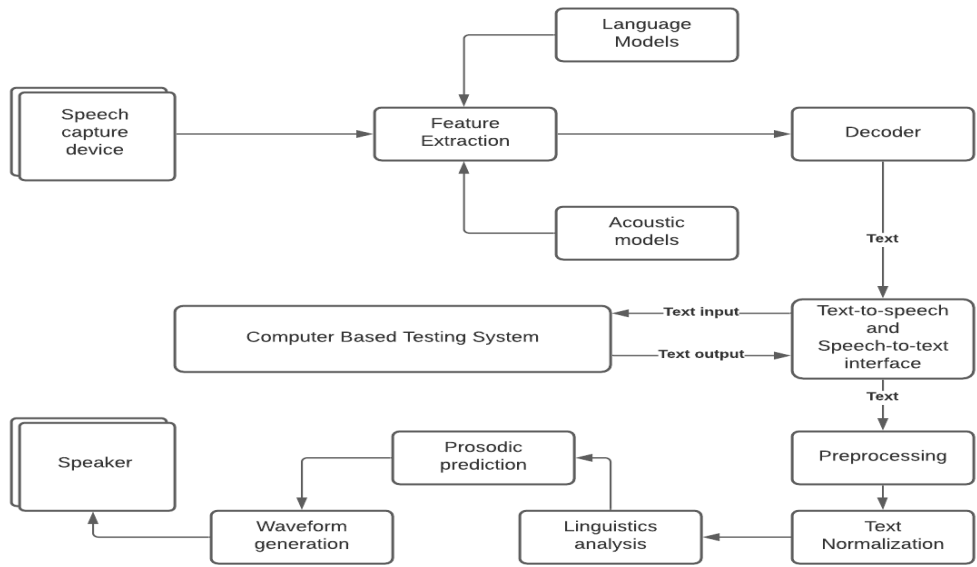

*Figure 1. Data Flow Diagram of the new System*

In Figure 1, each components of the design is further explained as follows;

**Speech capture device:** This tool is used to record spoken words, which can be either a microphone or voice recorder. This device is embedded in the NLP model that is responsible for the capturing of spoken words.

**Feature Extraction**: Feature extraction is a library in a machine learning algorithm that is responsible for converting raw text data into a set of relevant and meaningful features that can be used to represent the data for further analysis, modeling, and processing. It transforms raw and unstructured data into a structured format that can be understood and processed by machine learning algorithms. These extracted features then serve as input for training models or conducting various analyses.

**Language Models**: These are computational models designed to understand, generate, and manipulate human language. They process and generate text based on the patterns and structures it has learned from vast amounts of text data. They can comprehend and interpret text by extracting information, identifying entities, recognizing sentiment, and understanding context.

**Acoustic models:** It is a component in the automatic speech recognition (ASR) system, which converts spoken language into written text. These models specifically focus on analyzing and understanding the acoustic properties of speech signals, enabling the system to accurately transcribe spoken words.

**Decoder**: The decoder is responsible for generating the final output based on the information provided by the acoustic models and language models, it combines them and produces the final output. It enables the system to transform the predictions and possibilities from these models into a coherent and contextually accurate output.

**Text-to-speech and speech-to-text interface**: The final output gotten from the decoder which is in text format was converted into speech using the text-to-speech engine so as to generate a human-like voice that read out the text questions. While the speech-to-text interface converts the spoken words from the virtually impaired students into written text, both TTS and STT interfaces are responsible for making the CBT system more accessible and user-friendly.

**Preprocessing**: Preprocessing refers to a set of tasks and techniques applied to raw data before it is used for analysis, modeling, or other computational tasks. In NLP, preprocessing is essential in transforming and cleaning textual data into a format that can be effectively utilized by algorithms and models.

**Text Normalization**: It is a part of text preprocessing that involves transforming text into a standardized and consistent format. It is used to reduce variations in text data that can arise due to differences in spelling, grammar, punctuation, and other factors which can make it difficult for computers to accurately understand and analyze text.

**Linguistic Analysis**: It is the systematic examination and study of the structure, patterns, and components of language. It involves breaking down language into its various elements to understand how it works, how meaning is conveyed, and how communication is achieved. It provides the foundation for NLP algorithms to understand and generate human language.

**Prosodic prediction**: It refers to the patterns of pitch, rhythm, intonation, and stress in speech that convey various aspects of meaning and emotion beyond the individual words. Prosodic prediction involves estimating how these prosodic features will be applied to a given piece of text during speech synthesis or analysis. It enhanced the naturalness, expressiveness, and overall quality of synthesized speech. It ensures that the synthesized speech conveys the appropriate rhythms, intonation, and emotional nuances, making it sound more human-like and engaging.

**Waveform Generation**: This is to create an audio waveform that represents sound signals. In the context of TTS, waveform generation involves in converting text or phonetic information into audible speech which is very important in creating natural and human-like synthetic speech. Its goal is to create speech that not only sounds natural but also conveys appropriate rhythms, intonations, and emotional cues.

**Speaker**: The speaker produces sound or speech. It plays a crucial role in communication, audio reproduction, and technology, enabling the transmission of information and entertainment through spoken language and sound.

### 3.1 Architectural Design of the system

The architectural design of the NLP-guide CBT (see Figure 2) exam system consists of three layers, which are;

**User-friendly interface**: This creates an inclusive and user-friendly interface that seamlessly integrates screen reader capabilities, making the NLP-driven Computer-Based Test guide accessible to visually impaired students.

**Database**: This component is used to store the test questions, answer choices, user data, and other relevant information. A well-structured database is essential for storing and retrieving the necessary information, which facilitates the smooth functioning of the NLP-driven computer-based test for visually impaired students.

**NLP Module**: The NLP module is the core component of the system. It enables the system to understand and generate human language, facilitating communication between users and the application. The NLP module serves as the bridge between users and the computer-based test guide, enabling visually impaired students to interact naturally, access information, and receive relevant content in an accessible format.

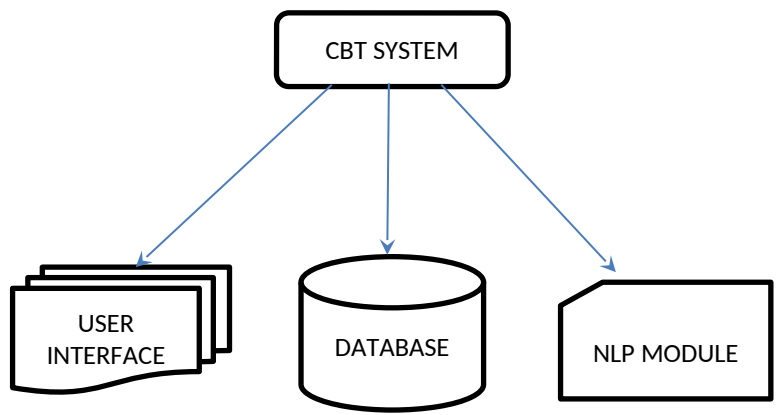

**Figure 2. Architectural Design of the System**

### 3.2. Dataset and Preparation

Creating an NLP-driven Computer-Based Test involves preparing relevant datasets to train and fine-tune the NLP model.

**Question and Answer data**: A diverse dataset of test questions, answer choices, and correct answers were gathered. This dataset was used to train the NLP model to understand and respond to user queries effectively.

**Synthesized speech data**: After the system has generated spoken responses, a synthesized datasets of speech recordings were collected, where these recordings cover various voices, tones, and speech rates, ensuring a variety of options for users.

**Fine-tuning data:** After training on the general NLP datasets, we used a smaller dataset containing examples specific to the test guide's context to fine-tune the model. This helped the model to adapt to the unique language and interaction patterns of the test guide.

**Train-validate-test split**: Here the datasets were divided into training, validation, and testing subsets. The training set was used to train the model, the validation set helped tune hyper-parameters, and the testing set evaluated the model's performance.

However, Audio datasets are collections of audio recordings that are used to train and evaluate machine learning models for tasks related to speech recognition and audio classification. This system was further tested with twenty (20) voice recordings from visually impaired students. Table 1. shows a sample of voice recordings of 5 visually impaired students.

**Table 1: A sample of 5 predicted audio datasets and the actual labels**

| RECORDS | RESPONSES | LABEL (A, B, C, D, E, F, G) |
|---|---|---|
| PERSON 1 | A | A |
| | B | B |
| | See | C |
| | D | D |
| | E | E |
| | F | F |
| | G | G |
| PERSON 2 | HEY | A |
| | BEE | B |
| | C | C |
| | D | D |
| | E | E |
| | F | F |
| | G | G |
| PERSON 3 | A | A |
| | B | B |
| | C | C |
| | D | D |
| | E | E |
| | F | F |

|  | G | G |
| --- | --- | --- |
| PERSON 4 | A | A |
|  | B | B |
|  | SEE | C |
|  | D | D |
|  | HE | E |
|  | F | F |
|  | G | G |
| PERSON 5 | A | A |
|  | A | B |
|  | C | C |
|  | D | D |
|  | HE | E |
|  | F | F |
|  | GEE | G |

Table 1 shows the system-generated speech and the actual label used to test the model. From the output responses generated by the system, it was observed that the system was able to capture the correct values spoken by the each student but in some cases couldn't get it right. This system is further evaluated to check the performance of the NLP mode for accuracy testing and also the values for precision, recall, and F1_score of the NLP model.

**Table 2: Precision, Recall, and F1-Measure**

| LABEL | TP | FP | FN | N | PRECISION | RECALL | F1_SCORE |
|---|---|---|---|---|---|---|---|
| A | 4 | 0 | 1 | 5 | 1.00 | 0.80 | 0.88 |
| B | 3 | 1 | 1 | 5 | 0.75 | 0.75 | 0.74 |
| C | 2 | 1 | 2 | 5 | 1.00 | 0.60 | 0.75 |
| D | 5 | 0 | 0 | 5 | 1.00 | 1.00 | 1.00 |
| E | 3 | 0 | 2 | 5 | 1.00 | 0.60 | 0.75 |
| F | 5 | 0 | 0 | 5 | 1.00 | 1.00 | 1.00 |
| G | 4 | 0 | 1 | 5 | 1.00 | 0.80 | 0.88 |

Table 2 explained the different columns like the "Truth Positive (TP)," "False Positive (FP)," "False Negative (FN)," "Number of students," "Label data," "Precision," "Recall" and "F1_score." The table shows the values obtained in each label and these values were used to calculate the precision, recall, and F1_score of each label showing the degree of accuracy of the NLP model.

Here we have a sample of 5 audio datasets and each of the labels was predicted to give the figures in TP, FP, and FN respectively.

The precision values of each of the labels indicate the quality of the speech generated by the system showing how some labels were predicted correctly while some were not exactly correct. Precision measures the accuracy of optimistic predictions.

Recall is a measure of how well a model correctly identifies True Positives. The recall values show the quantities of each of the labels that are predicted correctly.

F1-Score: It is a measure of a model's accuracy on the datasets. It combines both precision and recall of the model, and it is defined as the harmonic means of the model's precision and recall.

### 3.3. Experimental Tools

The system was developed using several tools and technologies such as Python, Flask framework, NLTK, PhpMyAdmin as the Relational Database Management System (RDMS), and Bootstrap. Python is a versatile, readable, extensive library support, and an open-source programming language. Flask framework is used to build an interactive web application. It is a library in Python that is both used for back-end and front-end development. NLTK is a Python library that is often used for text analysis and natural language processing tasks. PhpMyAdmin is a relational database management system that is used to interact with Python to store vital information on the CBT system. Bootstrap provides a collection of pre-designed HTML, CSS, and JavaScript components and tools that help to build an efficient, responsive, and visually appealing websites and web applications.

## 4. Result Discussion

This section discussed the performance of the NLP techniques for the NLP model applied on the datasets of labels (A, B, C, D, F, E, G) see (Table 2) where both the labels and the audio datasets were compared. The performance scores on the datasets were computed using precision, recall, and F1-score, see Table 3. Precision measures the accuracy of positive predictions. It focuses on the correctness of the positive predictions, while recall measures the ability of the model to find all the relevant instances. Then that of the F1_score measures the overall performance of the model which combines precision and recall into a single value, providing a balanced evaluation of the model's effectiveness.

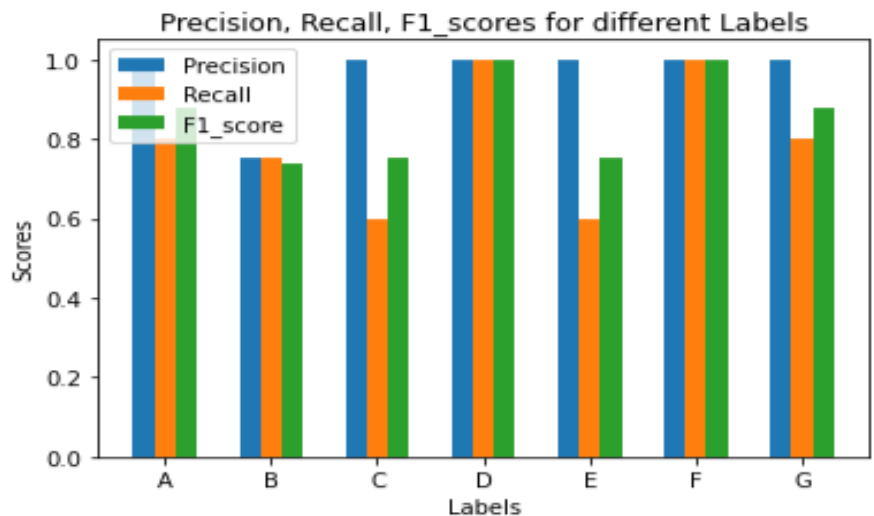

**Figure 3. Bar chart of precision, recall, and f1-score.**

The audio recordings shown in Table 4 were predicted and the outcomes show its level of accuracy and efficiency. The average predicted value for precision is 96%, that of recall is 79% while the F1_score value is 86%, showing that the NLP model is efficient enough to take in audio datasets and convert them into a written text. These metrics are used to assess the performance of the model and indicate its better version of the evaluated system. The Figures 4, 5, and 6 are the outputs of the CBT system with test questions and the options.

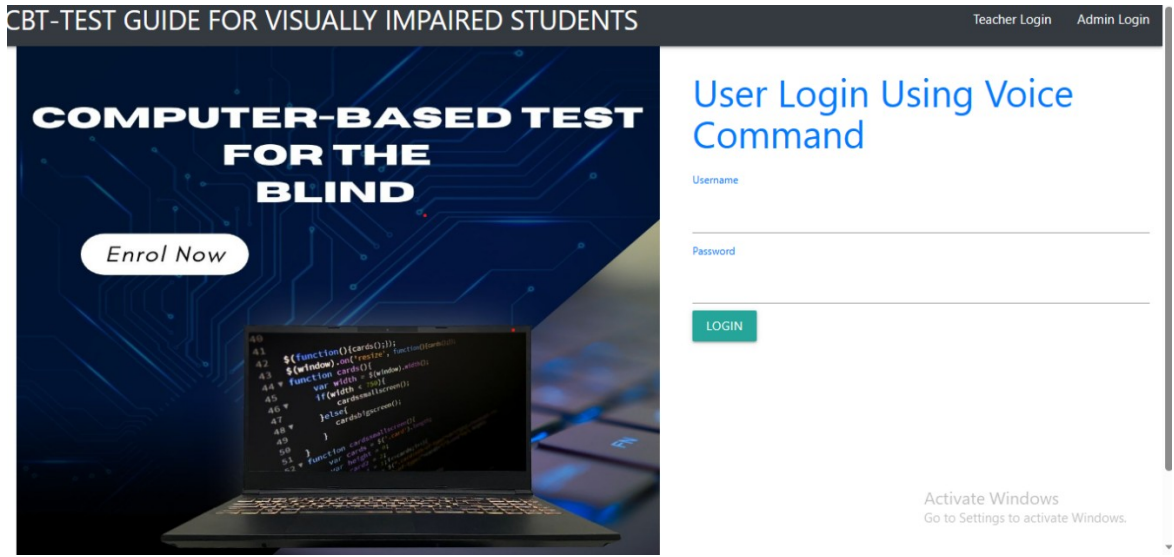

Figure 4. shows the login module with voice command

Figure 4 shows how the visually impaired student login to the system through voicing their login details and the system will confirmed the details with the details in the database, when matched it login the student .

Figure 5 shows the sample of the registered questions and their options.

```
Question 1  What is the Capital of England
A: London
B: Derby
C: Manchester
D: Nothingham Forest
Speak now...
Sorry, I didn't catch that.
Wrong!
Your score is 0
Question 2 Who is the richest man in the world
A: Bill Gate
B: Elon Musk
C: Bernard Arnault
D: Dangote
Speak now...
You said: a
Correct!
Your score is 1
Question 3 What is the addition of 5 + 6
A: 8
B: 11
C: 10
D: 12
Speak now...
You said: b
Correct!
Your score is 2
Question 4 The division of Nucleus is known as
A: karyokinesis
B: cytokinesis
C: isogamy
D: isopomy
Speak now...
You said: a
Correct!
Your score is 3
Question 5 The metal extracted from cassiterite is
A: Calcium
B: Copper
C: Tin
D: Sodium
Speak now...
You said: d
Correct!
```

Figure 6. Questions and answers section

Figure 6 shows how the system uses the Text-to-Speech library to read out the questions and options loud to the visually impaired student, then asked the student to speak now by allowing the student to speak to the microphone. When the student speaks, the system uses the Speech-to-Text library to capture the responses of the student and convert the spoken option to text. The system output the spoken word by telling the student the option he/she said and also confirmed if the answer was correct or not, thereafter the system automatically scores the student and reads out the scores result gotten by the student.

## 5. Conclusion

The NLP-driven Computer-Based Test guide is a promising solution to make computer-based testing more accessible and inclusive for visually impaired students. The guide will use NLP to classify different types of questions and provide navigation, text-to-speech, answer input, feedback, and customization features. The implementation of the guide will involve data collection, NLP model training, development of the user interface, integration with computer-based testing systems, testing, and refinement. The system's effectiveness has been demonstrated through successful trials and positive feedback from visually impaired students. However, this paper is not just concerned with the issue of CBT using TTS and speech recognition to read aloud text and also to capture voices but focuses on the application of NLP techniques to train a model. The results, it has shown that the system was capable of capturing voices correctly and converting those voices to text. This outcome is presented in a user-friendly interface for students to easily get to know the outcome of the CBT exercise using the Flask web framework for Python.